\documentclass[10pt, a4paper]{article}

\usepackage{natbib}
\usepackage{multibib}
\def\@mb@citenamelist{cite,citep,citet,citealp,citealt,citepalias,citetalias}
\newcites{languageresource}{~}

\usepackage{graphicx}
\usepackage{tabularx}

\usepackage{xcolor}
\usepackage{hyperref}

\usepackage{xstring}

\usepackage{color}
\usepackage{helvet}
\usepackage{url}
\usepackage{courier}
\usepackage{latexsym}
\usepackage{multirow}
\usepackage{xspace,mfirstuc,tabulary}
\usepackage{vcell}
\usepackage{colortbl}
\usepackage{amsmath}
\usepackage{booktabs} 
\usepackage{rotating}
\usepackage{pifont}

\usepackage{array}

\usepackage{lrec-coling2024} 

\title{ContrastWSD: Enhancing Metaphor Detection with Word Sense Disambiguation Following the Metaphor Identification Procedure}

\name{Mohamad Elzohbi, Richard Zhao} 
\address{University of Calgary \\
         Calgary, Alberta, Canada T2N 1N4 \\
         \{melzohbi,richard.zhao1\}@ucalgary.ca\\}

\abstract{
This paper presents ContrastWSD, a RoBERTa-based metaphor detection model that integrates the Metaphor Identification Procedure (MIP) and Word Sense Disambiguation (WSD) to extract and contrast the contextual meaning with the basic meaning of a word to determine whether it is used metaphorically in a sentence. By utilizing the word senses derived from a WSD model, our model enhances the metaphor detection process and outperforms other methods that rely solely on contextual embeddings or integrate only the basic definitions and other external knowledge. We evaluate our approach on various benchmark datasets and compare it with strong baselines, indicating the effectiveness in advancing metaphor detection. \\ \newline  }

\begin{document}

\maketitleabstract

\section{Introduction}

A metaphor, is a rhetorical device that compares, implicitly, two objects or concepts that are seemingly dissimilar but share symbolic or figurative similarities, with the intention of illuminating a fresh perspective and a more elaborate and nuanced comprehension of the world. Metaphors are not only intrinsic to creative writing, but they are also ubiquitous in human communication. Metaphors typically involve employing words in a manner that diverges from their basic definition, and their figurative sense is dependent on the context in which they are used. While novelty is an indicator of greater creativity in metaphors, sometimes they become widely used and established in the language, ultimately entering the lexicon as conventional metaphors, also known as dead metaphors \cite{lakoff2008metaphors}. 

\begin{figure}
	\centering
	\includegraphics[width=0.45\textwidth]{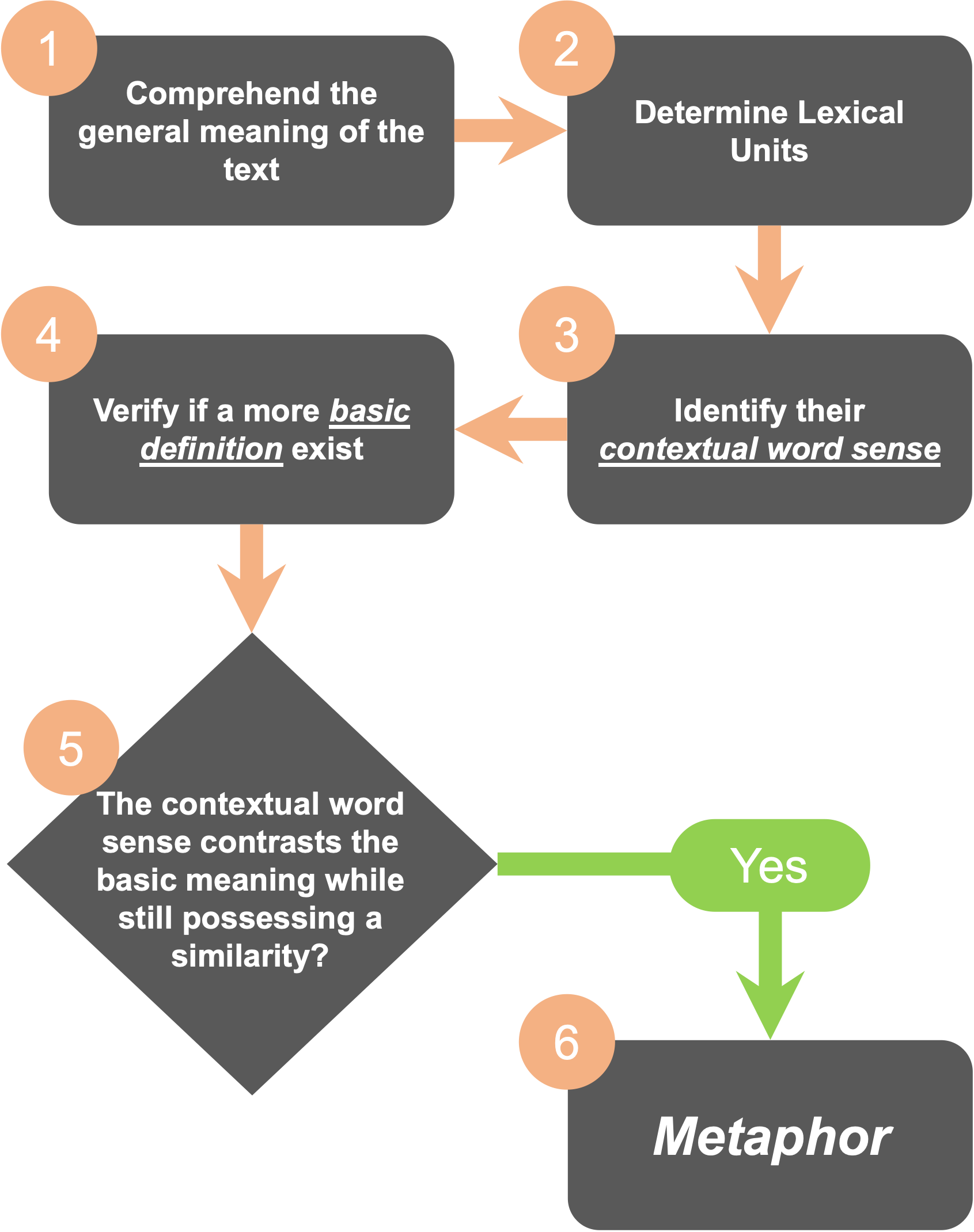}
	\caption{The Metaphor Identification Procedure}
	
	\label{fig:mip}
\end{figure}

Automatic metaphor detection, the process of identifying metaphoric expressions within a given text, is essential for various Natural Language Processing (NLP) tasks, such as sentiment analysis, text paraphrasing, and machine translation \cite{mao2018word}. The development of metaphor detection models presents a significant challenge, as it requires the identification and analysis of both the basic and contextual meanings of words within their respective contexts, as recommended by the Metaphor Identification Procedure (MIP) \cite{group2007mip, steen2010method} (see Figure \ref{fig:mip}).

Early approaches relied on extensive manual efforts in feature engineering. However, with the advent of word embedding techniques and neural networks, more efficient and effective methods emerged for this task \cite{song2021verb}. Notably, transformer-based models have demonstrated promising capabilities in detecting metaphors \cite{su-etal-2020-deepmet, choi2021melbert, babieno2022miss, li2023framebert}. Despite these advancements, there is still scope for further improvement to simulate the Metaphor Identification Procedure effectively. Therefore, the main objective of this study is to investigate the efficacy of transformer-based models in word sense disambiguation and in following the systematic Metaphor Identification Procedure to extract and contrast between the contextual word sense and the basic definitions of a target word to enhance automatic metaphor detection. 

Our proposed method is evaluated on established benchmark datasets, and the results demonstrate significant improvements. Comparatively, our model consistently achieves superior (and occasionally comparable) precision, recall, and F1 scores when compared to other recent and robust metaphor detection models.

\section{Related Work}

Metaphor detection has been a subject of active research in NLP for several years. Traditional approaches to metaphor detection have relied on hand-crafted or automatically acquired linguistic features, but recent advancements in NLP have resulted in the development of transformer-based pre-trained models that have demonstrated state-of-the-art performance across various NLP tasks \cite{choi2021melbert, song2021verb}.
DeepMet \cite{su-etal-2020-deepmet} formulates metaphor detection as a reading comprehension task and uses a RoBERTa-based model that incorporates global and local text context, question information, and part-of-speech as features to detect metaphors. 

MelBERT \cite{choi2021melbert} is a RoBERTa-based model that incorporates linguistic metaphor identification theories. MelBERT captures the context-dependent nature of metaphors and has demonstrated state-of-the-art performance on multiple benchmark datasets. While the authors considered the MIP procedure in their design, their focus was on leveraging contextual and out-of-context word embeddings to represent the word sense and basic definitions of the word. However, utilizing contextual word embeddings may not always accurately represent the word sense definition; instead, it may lean more towards the general contextual meaning. Similarly, out-of-context word embeddings may not necessarily reflect the basic meaning of the word, as they may be influenced by the frequent meaning, which might not align with the word's basic sense \cite{group2007mip, babieno2022miss}. In contrast, we encode both the contextual and basic definitions of the target words, which are extracted from the dictionary. This enables us to provide a more comprehensive understanding of their meanings and better align with the MIP procedure.

Researchers have also explored the use of external knowledge sources, such as definitions, word senses, and frames, to enhance the performance of metaphor detection models. For instance, \citeauthor{wan2021enhancing} (\citeyear{wan2021enhancing}) used gloss definitions to improve metaphor detection by considering both contextual embedding and contextual definition that best fits the context. Similarly, \citeauthor{babieno2022miss} (\citeyear{babieno2022miss}) explored the integration of the most basic definitions from Wiktionary to improve MelBERT's performance, achieving comparable or superior results. In contrast, our model extracts the contextualized definitions and contrasts them with the basic definitions to align with the MIP procedure. FrameBERT \cite{li2023framebert} proposed a new approach that incorporates FrameNet \cite{baker1998berkeley} embeddings to detect concept-level metaphors, achieving comparable or superior performance to existing models. Although encoding concepts may improve the model's understanding of similarities, it might not fully capture the variations in word meanings in various contexts. 

Word-Sense Disambiguation (WSD), which involves identifying the correct sense of a word in context, is a challenging task in NLP with various applications. Our study shows that WSD can aid in the process of identifying metaphors by disambiguating the word sense in given contexts. Multiple state-of-the-art WSD models have been proposed, including a modified version of BERT \cite{yap2020adapting} trained on a combination of gloss selection and example sentence classification objectives. \citeauthor{bevilacqua2020breaking} (\citeyear{bevilacqua2020breaking}) propose a method for incorporating knowledge graph information into WSD systems to improve their performance. The authors use a large-scale knowledge graph (DBpedia) to provide additional context and semantic information for each word in the text. SenseBERT \cite{levine2019sensebert} pre-trains BERT on a large-scale sense-annotated corpus using a modified loss function to incorporate sense-aware training objectives. 

Incorporating WSD models to obtain contextual definitions for use in metaphor detection has also been explored. For example, Metaphorical Polysemy Detection (MPD) \cite{hall2022metaphorical} has been proposed, which focuses on detecting conventional metaphors in WordNet. By combining MPD with a WSD model, this method can determine whether a target word represents a conventional metaphor within a given sentence. The authors have identified the two limitations mentioned earlier regarding MelBERT, which hinder its alignment with the MIP procedure. Particularly, attempting to implicitly infer word sense from the target word's contextual representation and assuming that the out-of-context embedding of the target word represents its basic meaning.

To address these issues, the authors trained an MPD model jointly with a WSD model to detect the metaphoricity of a target word, leveraging the word senses predicted by the WSD model for the target word in a given context. However, we argue that this model still lacks full alignment with MIP since it implicitly contrasts the basic definition with the word sense.

To achieve a more explicit alignment with MIP, we propose a different approach. Firstly, we utilize a WSD model to extract the word sense from a lexicon. Secondly, we tackle the second problem by considering the first definition listed in Wiktionary as the basic definition. This choice aligns with the dictionary's recommendation to utilize the logical hierarchy of word senses \cite{babieno2022miss}. The explicit contrast between the basic and word sense definitions corresponds better to steps 3 to 5 outlined in the flowchart of the MIP procedure (see Figure \ref{fig:mip}), as we elaborate on in the following section.

\begin{figure*}[htb]
	\centering
	\includegraphics[width=\textwidth]{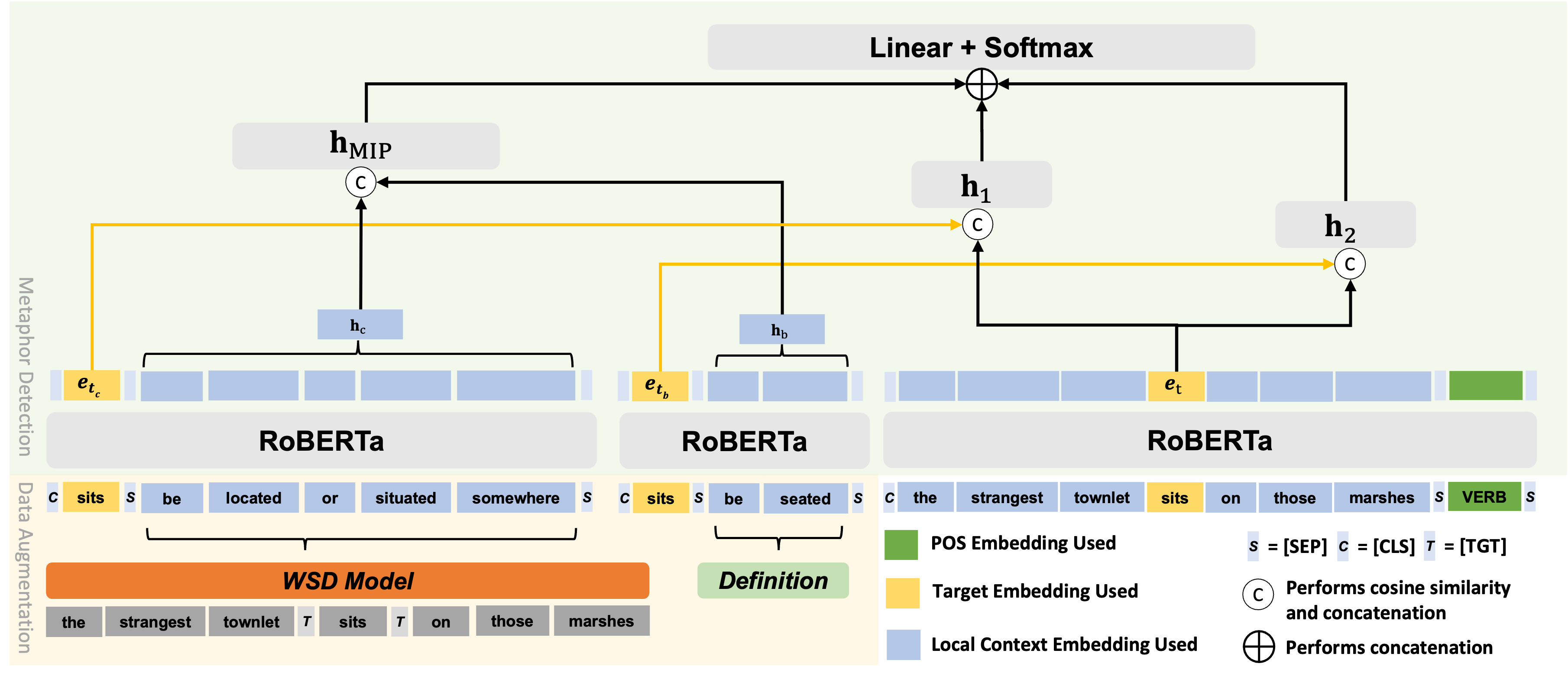}
	\caption{ContrastWSD overall framework showing both stages: (i) the data augmentation stage and (ii) the metaphor detection stage.}
	
	\label{fig:model}
\end{figure*}

\section{Methodology}
In this section, we present the methodology used to develop and train ContrastWSD. Figure \ref{fig:model} provides an illustration of the data augmentation process and the subsequent metaphor detection process. We commence by introducing the datasets utilized in the study, followed by an overview of the word-sense augmentation procedure. Subsequently, we outline the modifications that were made to the MelBERT model's structure to enhance metaphor detection.

\subsection{Data Augmentation}

A major contribution of our research is the adherence to the systematic approach of the Metaphor Identification Procedure for detecting linguistic metaphors in the VUA datasets. The MIP procedure (as outlined in Figure \ref{fig:mip}) involves: (1) comprehending the general meaning of the text, (2) determining lexical units, (3) identifying the contextual meaning of the units, (4) and verifying if there is a more basic meaning. (5) If the contextual meaning deviates from the basic meaning but remains understandable by comparison, (6) the unit is labeled as metaphorical. 

To align with MIP, we augmented the existing datasets used in the MelBERT model through a two-step procedure. Firstly, we employed a BERT WSD model fine-tuned on a sequence-pair ranking task \cite{yap2020adapting} to extract the word sense contextual definition of the target word from WordNet. To retrieve the contextual word sense, we feed a sentence $S_{c} = (w_1, ..., \texttt{[TGT]}, w_t, \texttt{[TGT]}, ..., w_n)$ to the WSD model, where $\texttt{[TGT]}$ is a special token marking the location of the target word $w_t$. The WSD model then performs gloss selection from WordNet and chooses the best definition $D_{c}$ of $w_t$ that fits the context. Secondly, we retrieved the basic definitions from the datasets compiled by \citeauthor{babieno2022miss} (\citeyear{babieno2022miss}). The authors selected the first definition listed in the Wiktionary dictionary as the basic definition, following the dictionary's recommendation to utilize the logical hierarchy of word senses in their guidelines. 

\subsection{Model Structure}
Our approach to metaphor detection, involves treating it as a binary classification problem based on the target word in a given sentence $S = (w_1, ..., w_t, ..., w_n)$. Our aim is to predict whether the target word $w_t$ is being used metaphorically or literally in the sentence. To accomplish this, we leverage the contextual and basic meanings of the target word in the given sentence. Our model utilizes three separate RoBERTa models to encode the sentence $S$, the isolated target word $w_t$, as well as the contextual and basic definitions $D_{c}$ and $D_{b}$. 

Following the MelBERT design, we modify the sentence $S$ by appending the POS tag to its end, and we enclose it with two segment separation tokens $\texttt{[SEP]}$. Additionally, we employ three types of extra embeddings: (1) The target embedding, used to indicate the target word. (2) The local context embedding, which marks either the words in the clause containing the target word between two commas or the definition and word sense. (3) The POS embedding, used to mark the position of the POS tag. We incorporate the target word and the definitions by prepending the word to the definitions we retrieve from WordNet (for the word sense) or Wiktionary (for the basic definition). This format is similar to how words appear in a dictionary, and we do this to utilize their hidden representation later in the model. We separate the target word from the definitions using the segment separation token $\texttt{[SEP]}$.

The RoBERTa models produce the hidden representations $\mathbf{h}_{S}$, $\mathbf{h}_{c}$, and $\mathbf{h}_{b}$ encoding the sentence, the contextual definition, and the basic definition, respectively with the extensions as described. $\mathbf{h}_{c}$ and $\mathbf{h}_{b}$ are produced by averaging the embedding of all the output tokens.
\begin{align*}
    (\mathbf{e}_{0}, ..., \mathbf{e}_{t}, ..., \mathbf{e}_{n}) &= \mathbf{h}_{S} 
    = \texttt{RoBERTa}(S) \\
    \mathbf{h}_{c} &= avg(\texttt{RoBERTa}(D_{c})) \\
    \mathbf{h}_{b} &= avg(\texttt{RoBERTa}(D_{b}))
\end{align*}

Since the MIP procedure focuses on the semantic contrast between the basic and contextual meanings of a word, we encode the basic and contextual meanings. Thus, our MIP layer uses the word sense embedding $\mathbf{h}_{c}$ and the basic definition embedding $\mathbf{h}_{b}$. We also use cosine similarity to measure the semantic gap between the embeddings, similar to the approach employed by \citeauthor{babieno2022miss} (\citeyear{babieno2022miss}).
\begin{equation}
\mathbf{h}_{MIP} = l( \mathbf{h}_{c} \oplus \mathbf{h}_{b} \oplus CosSim(\mathbf{h}_{c}, \mathbf{h}_{b}))
\end{equation}
Where $l(.)$ is a fully-connected layer. We have also introduced two additional helper layers to our model. The first layer learns the relationship between the target word's contextual embedding $\mathbf{e}_{t}$ and the target word embedding adjacent to the word's sense $\mathbf{e}_{t_c}$, while the second layer learns the relationship between the target word's contextual embedding $\mathbf{e}_{t}$ and the target word embedding adjacent to the word's basic definition $\mathbf{e}_{t_b}$. We believe that these helper layers will assist the MIP layer when the WSD model fails to distinguish between the word sense and the basic definition, particularly in the case of detecting novel metaphors that lack multiple definitions in the dictionary. 
\begin{align*}
    \mathbf{h}_{1} &= l(\mathbf{e}_{t_c} \oplus \mathbf{e}_{t} \oplus CosSim(\mathbf{e}_{t_c}, \mathbf{e}_{t})) \\
        \mathbf{h}_{2} &= l(\mathbf{e}_{t_b} \oplus \mathbf{e}_{t} \oplus CosSim(\mathbf{e}_{t_b}, \mathbf{e}_{t}))
\end{align*}
We concatenate the hidden vectors from the MIP and the two helper layers before feeding them to the final binary classification layer for metaphor prediction. 


\section{Experiments}
In this section, we present the datasets, baseline models, and experimental setup used to evaluate our model. We also discuss various aspects of our experimentation process.

\paragraph{Datasets:} Our study utilized the VU Amsterdam Metaphor Corpus (VUA) \cite{steen2010method} in five pre-processed versions: \textbf{VUA-18} \cite{leong2018report}, \textbf{VUA-20} \cite{leong2020report}, and \textbf{VUAverb}, which focuses exclusively on verb metaphors. Additionally, we used \textbf{VUA-MPD-All}, which is a subset that focuses on the words that are available in WordNet, and \textbf{VUA-MPD-Conv}, a subset that focuses on conventional metaphors \cite{hall2022metaphorical}. Each dataset comprises sentences with a labeled target word, categorized as either metaphorical or not, along with a corresponding POS tag for the target word. For each dataset, there are separate train and test datasets. The VUA-18 dataset further includes four testing subsets, each corresponding to a different POS: \textbf{verb, noun, adjective, and adverb}. We augmented the datasets with word senses extracted from WordNet using the BERT-based WSD model, and we included the basic definitions extracted from Wiktionary, as explained in the previous section.

However, we encountered instances where the WSD model failed to find a word sense or cases where there was no definition available in the dataset. To ensure a fair comparison between our model and the baselines, we created two versions of the VUA-18, VUA-20, and VUAverb training and testing datasets:

\begin{enumerate}
    \item The original datasets were retained without removing any records. In places where no word sense was produced, we substituted it with the target word itself. For these target words, our model behaves comparable to MelBERT, enabling a comparison of the target word's embedding within and outside the context. Only in this case, our model is dependent on the contextual embeddings of the target word without external knowledge. \label{itemone}
    \item We removed the records where the WSD model failed to find a word sense and marked the pruned dataset with a minus sign (-). This was done to avoid potential noise arising from incomplete information during the model's learning process and to enable fair comparisons, ensuring that the model trains only on instances where all the necessary information is available.
\end{enumerate}

For the VUA-MPD training and testing datasets, as well as the VUA-18 POS testing subsets, we used only one version which corresponds to how we handled the datasets in version \ref{itemone}.

\begin{table}[htb]
    \centering
    \resizebox{\columnwidth}{!}{

    \begin{tabular}{l|c|ccc}
        \toprule 
        Dataset & Model & Rec & Prec & F1 \\ \midrule
        \multirow{3}{4em}{VUA-18} & MelBERT & 77.5 & 79.87 & 78.66  \\ 
                                  & MsW\_cos & \underline{77.88} & \textbf{80.31} & \underline{79.07}  \\ 
                                  & FrameBERT & 76.78 & 79.33 & 78.03  \\
                                  \cmidrule(r){3-5}
                                  & ContrastWSD & \textbf{78.85} & \underline{80.16} & \textbf{79.50} \\ 
                                  \midrule
        \multirow{3}{5em}{VUA-18 (-)} & MelBERT & \underline{73.34} & 71.44 & 72.35  \\ 
                                  & MsW\_cos & 70.52 & \textbf{74.77} & \underline{72.55} \\ 
                                  & FrameBERT & 70.33 & \underline{73.81} & 72.02  \\
                                  \cmidrule(r){3-5}
                                  & ContrastWSD & \textbf{75.23} & 72.38 & \textbf{73.77} \\ \midrule

        \multirow{3}{4em}{VUA-20} & MelBERT & 69.51 & 75.58 & 72.40  \\ 
                                  & MsW\_cos & \textbf{69.98} & \underline{75.64} & \underline{72.70}  \\
                                  & FrameBERT & 69.30 & 75.62 & 72.31  \\
                                  \cmidrule(r){3-5}
                                  & ContrastWSD & \underline{69.89} & \textbf{76.60} & \textbf{73.09} \\ 
                                  \midrule
        \multirow{3}{5em}{VUA-20 (-)} & MelBERT & 74.61 & 71.26 & 72.88  \\ 
                                  & MsW\_cos & 77.22 & 68.39  & 72.51  \\ 
                                  & FrameBERT & 75.07 & 71.59 & 73.28  \\
                                  \cmidrule(r){3-5}
                                  & ContrastWSD & \textbf{75.57} & \textbf{72.95} & \textbf{74.22} \\ 
        \bottomrule
    \end{tabular}
    }
    \caption{Evaluation results on VUA-18, VUA-20, and their counter pruned datasets. A bold number corresponds to the best performing model, while the underlined number the second best.}
\label{table:1}
\end{table}

\begin{table}[htb]
    \centering
    \resizebox{\columnwidth}{!}{

    \begin{tabular}{l|c|ccc}
    \toprule 
        Dataset & Model & Rec & Prec & F1 \\ \midrule
        \multirow{3}{4em}{verb} & MelBERT & \underline{80.25} & 71.88 & 75.83  \\ 
                                  & MsW\_cos & \textbf{81.85} & 70.91 & \underline{75.97}  \\ 
                                  & FrameBERT & 75.11 & \underline{74.00} & 74.55  \\
                                  \cmidrule(r){3-5}
                                  & ContrastWSD & 78.87 & \textbf{75.70} & \textbf{77.25} \\ \midrule
        \multirow{3}{4em}{noun} & MelBERT & \underline{66.98} & 73.74 & (70.19)   \\ 
                                  & MsW\_cos & \textbf{68.54} & 73.12 & \underline{(70.76)}   \\ 
                                  & FrameBERT & 62.57 & \underline{75.35} & 68.35  \\
                                  \cmidrule(r){3-5}
                                  & ContrastWSD & 66.29 & \textbf{76.36} & \textbf{70.97}  \\ \midrule
        \multirow{3}{4em}{adverb} & MelBERT & \textbf{69.20} & 71.83 & \underline{70.43}   \\ 
                                  & MsW\_cos & 64.57 & 71.43 & 67.80   \\ 
                                  & FrameBERT & 65.90 & \underline{74.56} & 69.88  \\
                                  \cmidrule(r){3-5}
                                  & ContrastWSD & \underline{68.52} & \textbf{77.45} & \textbf{72.63}  \\ \midrule
        \multirow{3}{4em}{adjective} & MelBERT & \textbf{67.06} & 65.97 & \underline{66.47}   \\ 
                                  & MsW\_cos & \underline{66.35} & 64.39 & 65.34   \\ 
                                  & FrameBERT & 64.25 & \underline{68.09} & 66.06  \\

                                  \cmidrule(r){3-5}
                                  & ContrastWSD & 65.73 & \textbf{70.83} & \textbf{68.18}  \\
        \bottomrule
    \end{tabular}
    }
    \caption{Performance comparison by POS tags. The results in between brackets indicate no statistically significant differences compared to ContrastWSD.}
\label{table:2}
\end{table}

\paragraph{Baselines:} To evaluate the performance of our model, we conducted a comparative analysis with four other baseline models that share similar concepts. The first baseline is the original model, \textbf{MelBERT} \cite{choi2021melbert}. The second is a modified version of MelBERT called \textbf{MsW\_cos} \cite{babieno2022miss}, which incorporates the basic definitions. We specifically used the cosine similarity version, as it yielded mostly the best results, as reported in their paper. The third baseline model is \textbf{FrameBERT} \cite{li2023framebert}, which utilizes frame embeddings to detect concept-level metaphors. We applied these models to the VUA-18, VUA-20, and VUAverb datasets, as well as their corresponding pruned versions. Additionally, we evaluated these models on the POS testing subsets after training them on the VUA-18 training dataset. The fourth baseline model is \textbf{MPD\_WSD} \cite{hall2022metaphorical}, which trains a WSD model to detect conventional metaphors. As this model employs examples from WordNet and uses distinct subsets of the VUA dataset, which we refer to as (VUA-MPD), we ran our model on the same training and testing split.
\paragraph{Experimental Setup} The experimental setup involved using the VUA-18, VUA-20, and VUA-verb datasets, along with their corresponding pruned versions, which included training, testing, and validation subsets. For each experiment, our model and the baseline models (except the MPD\_WSD) were trained on the training datasets and then evaluated on the respective testing datasets (refer to Tables \ref{table:1} and \ref{table:3}). Additionally, the models trained on the VUA-18 training dataset were tested on the verb, noun, adverb, and adjective testing subsets (refer to Table \ref{table:2}), while the models trained on the VUA-20 training dataset were evaluated on the VUA-verb testing dataset which is appended by a star ($\star$) sign in Table \ref{table:3}. We also train and test our model on the VUA-MPD datasets (refer to Table \ref{table:4}). 

All models were trained for three epochs using a learning rate of $3e-5$, a linear scheduler with a two-epoch warmup, and a dropout ratio of 0.2. To ensure robustness, we repeated the experiments five times using seeds 1, 2, 3, 4, and 5, following the approach in \citeauthor{babieno2022miss} (\citeyear{babieno2022miss}). The evaluation results from the test datasets were averaged over the 5 runs. We set the metaphor class weight to 3 on the original datasets, while a class weight of 4 was used on the pruned datasets, addressing the imbalance within the dataset. The experiments were conducted on a 2-GPU NVIDIA Tesla V100 with 16GB memory. Our source code is available along with all the datasets on our GitHub page\footnote{\url{https://github.com/melzohbi/ContrastWSD}}. 

We presented the findings from the MPD\_WSD model as reported in the paper \cite{hall2022metaphorical} trained on both the VUA-MPD-All and VUA-MPD-Conv subsets. Unfortunately, we encountered memory constraints that prevented us from replicating the reported results. To maintain consistency with the MPD\_WSD model, which employs the BERT-base-cased model, we also developed an alternative version of our model trained on BERT-base-cased instead of RoBERTa-base appended with a ($\beta$) sign (see Table \ref{table:4}).

\section{Empirical Results and Case Studies}

\paragraph{Statistical Significance} The objective of this analysis is to assess the statistical significance of the performance improvements observed in the ContrastWSD model. We conduct a two-tailed t-test for each dataset, comparing our model to the baseline models, setting a significance level of p = 0.05. The results indicate that the reported differences between our model and the other baseline models are statistically significant at a confidence level of 95\%, with only a few exceptions. These exceptions, which are discussed later, are marked in brackets () in Tables \ref{table:2} and \ref{table:3}. We did not analyze the significance of performance of the MPD\_WSD compared to our model as we reported the results from their paper. In the tables, the results in \textbf{bold} correspond to the best performing model, while the \underline{underlined} results indicate the second best performing model.

\begin{table}[htb]
    \centering
    \resizebox{\columnwidth}{!}{
    \begin{tabular}{l|c|ccc}
    \toprule 
        Dataset & Model & Rec & Prec & F1 \\ \midrule
        \multirow{3}{4em}{VUAverb} & MelBERT & \textbf{81.08}  & 55.24 & 65.57  \\ 
                                  & MsW\_cos & 77.88 & 61.49 & 68.68  \\ 
                                  & FrameBERT & 73.33 & \textbf{71.95} & \textbf{(72.62)}  \\
                                  \cmidrule(r){3-5}
                                  & ContrastWSD & \underline{79.18} & \underline{66.97} & \underline{72.54}\\

                                  \midrule
        \multirow{3}{6em}{VUAverb (-)} & MelBERT & \textbf{81.40} & 51.27 & 62.87  \\ 
                                  & MsW\_cos & 79.26 & 59.46 & 67.92  \\ 
                                  & FrameBERT & 74.63 & \textbf{70.68} & \textbf{(72.56)}  \\

                                  \cmidrule(r){3-5}
                                  & ContrastWSD & \underline{79.28} & \underline{66.66} & \underline{72.42}\\ 
                                  \midrule
        \multirow{3}{6em}{VUAverb ($\star$)} & MelBERT & 72.22 & 76.45 & 74.27  \\ 
                                  & MsW\_cos & \textbf{75.09} & \underline{78.00} & \textbf{(76.51)}  \\ 
                                  & FrameBERT & 69.96 & 77.60 & 73.55  \\

                                  \cmidrule(r){3-5}
                                  & ContrastWSD & \underline{73.81} & \textbf{78.39} & \underline{76.03}\\ 

        \bottomrule
    \end{tabular}
    }
    \caption{Evaluation results on VUA-verb, VUA-verb (-), and on the VUA-verb ($\star$) datasets.}
\label{table:3}
\end{table}

\begin{table}[!htb]
    \centering
    \begin{tabular}{c|c|c}
    \toprule 
        Model & F1 - All & F1 - Conv  \\ \midrule
        MPD\_WSD & 63.10 &  65.90 \\ 
        ContrastWSD ($\beta$) & \underline{73.42} & \underline{72.31} \\
        ContrastWSD & \textbf{73.84} & \textbf{73.07} \\ \bottomrule
    \end{tabular}
    \caption{Performance comparison on the VUA-MPD-All and the VUA-MPD-Conv subsets.}
\label{table:4}
\end{table}

\paragraph{Overall Results} 

Table \ref{table:1} presents a comparison of evaluation results for the VUA-18, VUA-20, VUA-18 (-), and VUA-20 (-) datasets. 
The ContrastWSD model consistently outperforms all baseline models in terms of F1-score on both VUA-18 and VUA-20 datasets, even though these datasets had instances where the word sense was absent. Interestingly, the improvement in F1-score for ContrastWSD compared to other models doubled on the VUA-18 (-) and VUA-20 (-) datasets. This suggests that while the presence of word sense in the dataset may have contributed to the model's superior performance, its absence, occasionally, did not hinder its overall effectiveness.

Table \ref{table:2} presents the performance comparison for the verbs, adjectives, nouns, and adverbs testing subsets. Our model shows a notable advantage in detecting metaphorical adverbs, surpassing the other models by at least $2\%$ in F1-score. Additionally, it achieves at least a $1\%$ improvement on verbs and adjectives. Furthermore, it significantly outperforms the recent FrameBERT model by more than $2\%$ on the noun dataset, while achieving comparable results to MsW\_cos and MelBERT with at least a $0.21\%$ increase in F1 that was not statistically significant, while showing a notable precision gain. 

Table \ref{table:3} presents the comparison between our model and the baseline models on the VUA-verb and VUA-verb (-) datasets. The results demonstrate that our model significantly outperformed MelBERT and MsW\_cos on the small training dataset. Moreover, when the missing word senses were removed, our model's performance remained consistently better compared to MelBERT and MsW\_cos. On the other hand, while FrameBERT showed an insignificant gain of performance than our model on the small training datasets, its performance significantly declined when tested with the model trained on the larger VUA-20 training dataset. This observation suggests that our model maintains good performance even with small datasets, showing no signs of overfitting or noise introduction when trained on a larger dataset. Additionally, the performance of our model improved with considerably higher precision when trained on the larger dataset.

Table \ref{table:4} compares the performance on the VUA-MPD subsets used by the MPD\_WSD model. We compare the F1-score reported in their paper against our model's results, revealing a significant improvement of a minimum of $10.32\%$ F1-score gain for our models on the subset encompassing both novel and conventional metaphors, as well as a minimum of $6.41\%$ F1-score gain in detecting conventional metaphors.

\begin{table*}[!htb]
    \centering
    \begin{tabular}{ccccc|p{40mm}|p{30mm}|p{40mm}}
    \toprule 

        \begin{turn}{90}\textbf{True Label}\end{turn} & \begin{turn}{90}\textbf{ContrastWSD}\end{turn} & \begin{turn}{90}\textbf{FrameBERT}\end{turn} & \begin{turn}{90}\textbf{MsW\_cos}\end{turn} & \begin{turn}{90}\textbf{MelBERT}\end{turn} & \multicolumn{1}{c|}{\multirow{2}{30mm}{\textbf{Sentence Context} \\}} & \multicolumn{1}{c|}{\textbf{Word Sense}} & \multicolumn{1}{c}{\textbf{Basic Definition}} \\ 
        \midrule
        \ding{51} & \ding{51} & \ding{53} & \ding{53} & \ding{53} & ... and pull all nuclear \textbf{\textcolor{blue}{plant}} out of the impending sale ... & Buildings for carrying on industrial labor. & An organism that is not an animal, especially an organism capable of photosynthesis. \\
        \midrule
        \ding{53} & \ding{53} & \ding{53} & \ding{53} & \ding{53} & It is a good time to \textbf{\textcolor{blue}{plant}} hardy shrubs too. & (botany) a living organism lacking the power of locomotion. & An organism that is not an animal, especially an organism capable of photosynthesis. \\
        \midrule
        \ding{51} & \ding{51} & \ding{51} & \ding{53} & \ding{53} & ... 1,500-tonne consignment of Canadian PCBs (polychlorinated biphenyls) bound for this \textbf{\textcolor{blue}{plant}} in Gwent. & Buildings for carrying on industrial labor. & An organism that is not an animal, especially an organism capable of photosynthesis. \\
        \midrule
        \ding{51} & \ding{51} & \ding{53} & \ding{53} & \ding{53} & Stars appear and the shadows are fallin'. You can hear my \textbf{\textcolor{blue}{honey}} callin' & A beloved person; used as terms of endearment. & A viscous, sweet fluid produced from plant nectar by bees ... \\
        \midrule
        \ding{51} & \ding{51} & \ding{51} & \ding{51} & \ding{51} & ... but the tops of the mountains are still golden, as though \textbf{\textcolor{blue}{honey}} had been poured lightly over them ... & A sweet yellow liquid produced by bees. & A viscous, sweet fluid produced from plant nectar by bees ... \\
        \midrule

        \ding{53} & \ding{53} & \ding{51} & \ding{51} & \ding{51} & ... I'll be \textbf{\textcolor{blue}{jumping}} up and down like a what d' ya call it! & Move forward by leaps and bounds. & To propel oneself rapidly upward, downward and/or in any horizontal direction ... \\
        \midrule
        \ding{51} & \ding{51} & \ding{51} & \ding{51} & \ding{51} & ... there are plenty of girls who would \textbf{\textcolor{blue}{jump}} at the chance. & Enter eagerly into. & To propel oneself rapidly upward, downward and/or in any horizontal direction ... \\
        
    \bottomrule
    \end{tabular}
    \caption{Examples of predictions made by ContrastWSD and the baseline models on the VUA-20 testing dataset. \ding{51} marks a metaphor prediction and \ding{53} marks a literal prediction.}
\label{table:5}
\end{table*}

\paragraph{Case Studies:} As shown in the results, our ContrastWSD model exhibits relatively higher gains. To exemplify instances where ContrastWSD correctly labeled examples that were incorrectly labeled by the baselines, we conducted several case studies. Table \ref{table:5} presents a few cases that demonstrate the benefits of our approach, involving contrasting word senses while considering the context of the target sentence. For these case studies, we selected the models trained on the VUA-20 dataset. For each model, we chose the best performing one among the 5 seeds. The examples shown in the table are drawn from the VUA-20 testing dataset.

For instance, we observed the word ``plant'' mentioned 15 times in the testing dataset: 14 times in a metaphorical sense and only once in a non-metaphorical sense. Both MsW\_cos and MelBERT labeled all of these occurrences as non-metaphorical. In contrast, our model correctly identified $47\%$ of these occurrences, while FrameBERT only identified $33\%$ correctly. The other models have only recognized the non-metaphorical instance and none of the metaphorical instances correctly. Three of these examples are mentioned in the table.

Another example involves the word ``honey'', which was mentioned twice as a metaphor in the testing dataset. In the first instance, it was used in a conventional way, and our model correctly annotated it as metaphorical, leveraging the extracted word sense. The other models did not recognize this metaphorical usage. In the second instance, ``honey'' appeared as a novel metaphor where our model, along with the other baseline models, marked it as metaphorical. Even though the word sense was similar to the basic sense, our model still identified it as metaphorical. This indicates that our model can recognize both novel and conventional metaphors.

Finally, the word ``jump'' occurs two times in the testing dataset, appearing in two different tenses and senses. In one instance, the word ``jumping'' was used in the literal sense, and our model correctly identified it as literal, considering that the word sense was similar to the definition. However, the other models did not recognize it as such. In the other occurrence, ``jump'' was used metaphorically, and our model, along with the other models, correctly identified it as metaphorical.

\section{Conclusion and Future Work}

In this paper, we presented a RoBERTa-based model for metaphor detection that follows the Metaphor Identification Procedure by utilizing a WSD model to extract and contrast the contextual meaning with the basic meaning of a target word. We evaluated our model on several benchmark datasets and demonstrated that leveraging senses and contrasting them can enhance the performance of metaphor detection models. Our proposed model outperformed other state-of-the-art metaphor detection models. Our work provides compelling evidence for further exploration of the use of WSD models and sense-contrasting techniques to enhance the performance of metaphor detection models. 

In future work, we plan to investigate the integration of commonsense models such as COMET \cite{bosselut2019comet} to extract and utilize the common sense knowledge from the target word. COMET was used extensively in recent metaphor generation models \cite{elzohbicreative}. We believe that this integration will enable better differentiation of novel metaphors from nonsensical expressions.


\nocite{*}
\section{Bibliographical References}\label{sec:reference}

\bibliographystyle{lrec-coling2024-natbib}
\bibliography{lrec-coling2024-example}


\end{document}